\documentclass[sigconf]{acmart}

\usepackage{hyperref}
\usepackage{graphicx}
\usepackage{url}
\usepackage{xspace}
\usepackage{booktabs} 
\usepackage{multirow}

\newcommand\HumanAI{{\small\textsc{Human-AI}}\xspace}

\newcommand\sect[1]{\S\ref{#1}}

\AtBeginDocument{%
  \providecommand\BibTeX{{%
    \normalfont B\kern-0.5em{\scshape i\kern-0.25em b}\kern-0.8em\TeX}}}

\copyrightyear{2021}
\acmYear{2021}
\setcopyright{acmcopyright}\acmConference[FAccT '21]{Conference on Fairness, Accountability, and Transparency}{March 3--10, 2021}{Virtual Event, Canada}
\acmBooktitle{Conference on Fairness, Accountability, and Transparency (FAccT '21), March 3--10, 2021, Virtual Event, Canada}
\acmPrice{15.00}
\acmDOI{10.1145/3442188.3445923}
\acmISBN{978-1-4503-8309-7/21/03}

\begin{document}

% \title{[tentative] Formalizing Trust in Artificial Intelligence \\ (or) What is Trust in Artificial Intelligence? \\ (or) ?}
\title[Formalizing Trust in Artificial Intelligence: Prerequisites, Causes and Goals of Human Trust in AI]{Formalizing Trust in Artificial Intelligence: \\ Prerequisites, Causes and Goals of Human Trust in AI}

 \author{Alon Jacovi}
\affiliation{%
  \institution{Bar Ilan University}
}
\email{alonjacovi@gmail.com}

\author{Ana Marasovi\'{c}}
\affiliation{%
  \institution{Allen Institute for Artificial Intelligence}
  \institution{University of Washington}
  }
\email{anam@allenai.org }
% \email{alon}

\author{Tim Miller}
\affiliation{%
  \institution{School of Computing and Information Systems\\The University of Melbourne}
}
\email{tmiller@unimelb.edu.au}

\author{Yoav Goldberg}
\affiliation{%
 \institution{Bar Ilan University}
 \institution{Allen Institute for Artificial Intelligence}
 }

\email{yoav.goldberg@gmail.com}

\renewcommand{\shortauthors}{Alon Jacovi, Ana Marasovi\'{c}, Tim Miller, Yoav Goldberg.}

\begin{abstract}
Trust is a central component of the interaction between people and AI, in that `incorrect' levels of trust may cause misuse, abuse or disuse of the technology. But what, precisely, is the nature of trust in AI? What are the prerequisites and goals of the cognitive mechanism of trust, and how can we promote them, or assess whether they are being satisfied in a given interaction? This work aims to answer these questions. We discuss a model of trust inspired by, but not identical to, interpersonal trust (i.e., trust between people) as defined by sociologists. This model rests on two key properties: the \textit{vulnerability} of the user; and the ability to \textit{anticipate} the impact of the AI model's decisions. We incorporate a formalization of `contractual trust', such that trust between a user and an AI model is trust that some implicit or explicit contract will hold, and a formalization of `trustworthiness' (that detaches from the notion of trustworthiness in sociology), and with it concepts of `warranted' and `unwarranted' trust. We present the possible causes of warranted trust as intrinsic reasoning and extrinsic behavior, and discuss how to design trustworthy AI, how to evaluate whether trust has manifested, and whether it is warranted. Finally, we elucidate the connection between trust and XAI using our formalization.
\end{abstract}





\maketitle

\section{Introduction}

%\ana{I formatted paragraph titles in Section 3.2. in bold, let me know if that looks better. I can change in the entire draft.} \todo{yes, it looks good (though i hope the formatting allows it). i tried overwriting the paragraph command earlier to bold itself but recursive overwriting caused it to hang :)}

With the rise of opaque and poorly-understood machine learning models in the field of AI, trust is often cited as a key desirable property of the interaction between any user and AI \cite{das2020opportunities-xai,tjoa2019xai-survey,xu2019xai-history,carvalho2019interpretability-survey}. The recent rapid growth in \textit{explainable AI} (XAI) is also, in part, motivated by the need to maintain trust between the human user and AI  \citep{lipton2016mythos,kim2017interpretability,miller2017social,Ribera2019CanWD,toreini2020trust-trustworthiness,jacovi2020faithfulness}. By designing AI that users can and will trust to interact with,  AI can be safely implemented in society. 
% \footnote{\todo{https://ico.org.uk/about-the-ico/research-and-reports/project-explain-interim-report/ also applies here, but not sure how to cite it}}

However, literature seldom discusses specific models of trust between humans and AI. What, precisely, are the prerequisites for human trust in AI? For what goals does the cognitive mechanism of trust exist? How can we design AI that facilitates these prerequisites and goals? And how can we assess whether the prerequisites exist, and whether the purpose behind the trust has been achieved?

In this work, we are interested in formalizing the `trust' transaction between the user and AI, and using this formalization to further our understanding of the requirements behind AI that can be integrated in society. We consider `artificial intelligence' to be any automation that is attributed with intent by the user \cite[social attribution,][]{miller2017social}, i.e., anthropomorphized with a human-like reasoning process. For our purpose, we consider the user to be an individual person, rather than an organization, though aspects of the work are applicable to the latter as well. 

There are many vague aspects of trust that are difficult to formalize with the tools available to us in literature on AI and Human-Computer Interaction (HCI). For this reason, we first discuss how interpersonal trust is defined in sociology, and derive a basic, yet functional, definition of trust between a human and an AI model, based on the \textit{prerequisites} and \textit{goals} of the trustor gaining trust in the AI (Section \ref{sec:trust-basic-definition}). Specifically, the trustor must be \textit{vulnerable} to the agent's actions, and the trustor's goal in developing trust is to \textit{anticipate} the impact of the AI model's decisions. 
%By ``anticipate'', we mean that the user believes the AI will work `as intended'.
% \footnote{``Predicting'' the AI actions means that the user has a set of expectations regarding the desired behavior of the AI, and the user believes that these expectations will be met with high probability. This is explored later in the work.} This definition forms the foundation for the rest of the work.

However, the above definition is incomplete: though the goal is anticipating `intended' behavior, what can we say about when and whether this goal is achieved? We develop the definition further by answering two questions: (1) \textit{what is the AI model being trusted with (i.e., what is `intended')?}; and (2) \textit{what differentiates trust that achieves this goal, and trust that does not?} Section \ref{sec:contractual-trust} answers (1) via a notion of \textit{contractual trust}, and Section \ref{sec:trust-trustworthy} answers (2) via notions of \textit{warranted} and \textit{unwarranted} trust. In Section \ref{sec:human_ai_definitions} we complete the definition of \HumanAI trust with a formal summary of the above.

With these definitions, we are now equipped to discuss the \textit{causes} of trust in the AI (specifically, warranted trust in a particular contract), and how we should pursue the development of AI that will be trusted. In Section \ref{sec:causes-of-trust}, we answer the question: \textit{what are the mechanisms by which an AI model gains the trust of a person?} Namely, we define and formalize notions of \textit{intrinsic trust}, which is based on the AI's observable reasoning process, and \textit{extrinsic trust}, which is based on the AI's external behavior. 

Both intrinsic and extrinsic trust are deeply related to XAI. As mentioned, the XAI literature frequently notes trust as a principal motivation in the development of explanations and interpretations in AI, but seldom elucidates the precise connection between the methods and the goal. In Section \ref{sec:explainability-trust}, we unravel this `goal' of XAI ---to facilitate trust---by using our formulation thus far.

In Section \ref{sec:evaluation} we pivot to the question of evaluating trust, by discussing the evaluation of the vulnerability in the interaction, and of the ability to anticipate. Finally, in Section \ref{sec:discussion} we expand on other aspects of interpersonal trust and human-machine trust (automation not attributed with intent), their relation to our notion of \HumanAI trust, and 
possible future extensions of our formalization. 

\noindentparagraph{\textbf{Contributions.}} We provide a formal perspective of \HumanAI trust that is rooted in, but nevertheless not the same as,  interpersonal trust as defined by sociologists. We use this formalization to inform notions of the causes behind \HumanAI trust, the connection between trust and XAI, and the evaluation of trust. We hope that this work enables a principled approach to developing AI that should, and will, be trusted in practice.

% \begin{enumerate}
%     \item Is the trust of a person in an AI model different from, or the same as, trust between people? How can we leverage research on the latter to pursue the former? (\S\ref{sec:trust-basic-definition}, \S\ref{sec:contractual-trust}, \S\ref{sec:trust-trustworthy})
%     \item What exactly is an AI model being trusted with? Is it always the same thing? (\S\ref{sec:contractual-trust})
%     \item Is trust a singular desirable trait in the interaction, or are there different varieties of trust? If different kinds of trust exist, are all of them desirable as goals of AI development and XAI? (\S\ref{sec:trust-trustworthy})
%     \item What are the mechanisms by which an AI model gains the trust of a person? Is it identical or different in some way from interpersonal trust? (\S\ref{sec:causes-of-trust})
%     \item What should the evaluation of trust satisfy? Do current methods of evaluating trust fulfill these requirements? (\S\ref{sec:evaluation})
% \end{enumerate}

% Despite trust being a common motivator in AI development, little is currently known about any of these questions. We aim to answer these questions in this work.

% \tm{If I understand correctly, we are only talking about trust at an individual level, rather than at a societal level. That is, the trust that a user builds over time interacting with a machine, but not the trust that a particular organisation may have that a machine is fit for purpose. Perhaps we should scope this here in the introduction}

%\noindentparagraph{Organization of the work.}
\noindentparagraph{\textbf{Note on the organization of the work.}} The following sections provide an informal description of trust in AI via a narrative, in the interest of accessibility (\S\ref{sec:trust-basic-definition},\ref{sec:contractual-trust},\ref{sec:trust-trustworthy}). We provide formal, concise definitions of our taxonomy \textit{after} completing the relevant explanations (\S\ref{sec:human_ai_definitions}). Additionally, for coherency we bypass some nuance behind our choice of formalization, made available in \S\ref{sec:discussion}.

\section{A Basic Definition of Trust} \label{sec:trust-basic-definition}

To understand human trust in AI (\HumanAI trust), a useful place to start is to examine research in philosophy, psychology, and sociology of how people trust each other (\textit{interpersonal} trust). In this section, we present a primitive (and incomplete, as we will show) definition of trust that will serve as a basis for the rest of the work. 

%\noindentparagraph{Definition (\textit{interpersonal trust}).}
\noindentparagraph{\textbf{Definition (\textit{Interpersonal trust}).}} A common basic definition of trust regards it as a directional transaction between two parties: if A believes that B will act in A’s best interest, and accepts vulnerability to B’s actions, then A trusts B \citep{mayer1995integrative}. The goal of trust is to ``make social life predictable [by anticipating the impact of behavior], and make it easier to collaborate between people'' \citep{misztal1996trust}.\footnote{This definition of trust is considered overly simplistic by many in sociology. In Section \ref{sec:discussion} we discuss aspects of more elaborate formalizations of interpersonal trust, and whether they are relevant to \textsc{Human-AI}\xspace trust.}

Noteworthy in this definition, and key to defining \HumanAI trust, are the notions of \textit{anticipation} and \textit{vulnerability}. In particular, interpersonal trust exists to mitigate uncertainty and risk of collaboration by enabling the trustor's ability to anticipate the trustee---where `anticipating' refers to a belief that the trustee will act in the trustor's best interests. We maintain that \HumanAI trust exists for the same purpose, as a sub-case of trust in automation, following \citet{hoffman2017taxonomy-trust-automation}: trust is an attempt to anticipate the impact of behavior under risk.
% \noindentparagraph{Definition (\textit{\HumanAI trust}).} Trust is a directional transaction between two parties: A (human) and B (AI), where an AI is any machine whose behavior is anthropomorphized by the human. If A believes that B will act in A’s best interest, and accepts vulnerability to B’s actions, then A trusts B. The goal of this trust is to enable A to collaborate with B in the presence of A's risk.
Based on this, we conclude:

%\noindentparagraph{\textit{Risk} is a necessary prerequisite to the existence of \HumanAI trust.} 
\noindentparagraph{\textbf{\textit{Risk} is a %necessary
prerequisite to the existence of \HumanAI trust.}} %In this work,
We refer to risk as a disadvantageous or otherwise undesirable event to the trustor (that is a result of interacting with the trustee), which can possibly---\textit{but not certainly}---occur \citep{stanford-risk}. Therefore, ``to act in A's best interest'' is to avoid any unfavorable events. Admitting vulnerability means that the trustor perceives both of the following: (1) that the event is undesirable; and (2) that it is possible. Ideally, the existence of trust can only be verified after verifying the existence of risk, i.e., by proving that both conditions hold.  

% For example, receiving advice from AI in a trivia game \citep{feng-shi2019-triviabowl} involves the risk of the advice being incorrect, and losing as a result. The loss event must be undesirable to the user, and the user must understand that the advice could theoretically be incorrect, and that it is not certainly incorrect, for trust to manifest. \todo{not confident that this is a good example. but the other obvious examples like self-driving cars are inappropriate...}
% \tm{Yes, this is a bit too low stakes to be interest. What about the canonical example of credit scoring: if the loan officer makes the wrong decision either they risk the applicant defaulting in future, or they miss out on income (or more accurately, they pass on this but give it to a future applicant who is less credit worthy). For even higher stakes, consider the COMPAS algorithm for predicting prisoner re-offending, used by judges in parole decisions.}

For example, AI-produced credit scoring \cite{chen2018fico-credit} represents a risk to the loan officer: a wrong decision carries a risk (among others) that the applicant defaults in the future. The loss event must be undesirable to the user (the loan officer), who must understand that the decision (credit score) could theoretically (and not certainly) be incorrect for trust to manifest.
Similarly, from the side of the applicants (if they have a choice as to whether to use the AI model), the associated risk is to be denied or to be charged a higher interest rate on a loan that they deserve, and trust manifests if they believe that the AI model will work in their interest (the risk will not occur).
%\yg{Discuss also trust from the side of the person who is being evaluated? which is associated with a different risk.} \todo{on the simplest level of the example the applicant is not a user, so i think expanding the example to one where he is is difficult/confusing. i think we can accomplish the same just by adding a note/footnote about being thoughtful about who we're designating as the user in the definition, and that the same AI use-case can have multiple interactions with different users (the developer, the deployer, the customer, etc). I also think we're fine not doing that and keeping this simple here, because i tried to keep this section light intentionally to make the rest of the paper less brain intensive}

\noindentparagraph{\textbf{Distrust manifests in attempt to mitigate the risk.}} The notion of distrust is important, as it is the mechanism by which the user attempts to avoid the unfavorable outcome. We adapt \citeauthor{tallant2017contractual}'s definition of distrust: A distrusts B if A does not accept vulnerability to B's actions, because A believes that B may not act in A's best interest \cite{tallant2017contractual}. Importantly, distrust is \textit{not} equivalent to the absence of trust  \cite{stanford-trust}, as the former includes some belief, where the latter is lack of belief---or in other words, distrust is trust in the negative scenario. For the remainder of this paper, we focus our analysis on trust, as the link to distrust is straightforward.

%\noindentparagraph{\textit{Predictability} is a goal, but not necessarily a symptom, of \HumanAI trust.} 
\noindentparagraph{\textbf{The ability to anticipate is a goal, but not necessarily a symptom, of \HumanAI trust.}} The ability or inability of the user to anticipate the behavior of an AI model in the presence of uncertainty or risk, is \textit{not} indicative of the existence or absence of trust. We illustrate this in \sect{sec:trust-trustworthy}. We stress that anticipating intended behavior is the \textit{user's} goal in developing trust, but not necessarily the \textit{AI developer's} goal. 

%\todo{i don't think predictability needs an example because it's expanded (with examples) in the next 2 sections}

\section{Contractual Trust} \label{sec:contractual-trust}

The above notion of anticipating ability is incomplete. If the goal of trust is to enable the trustor's ability to anticipate, \textit{what} does the human trustor anticipate in the AI's behavior?
% (note that in the credit-scoring example, two users had somewhat different expectations from the system)? 
And what is the role of the `anticipated behavior' in the definition of \HumanAI trust?

\subsection{Trust in Model Correctness}
\label{sec:trust_model_correctness}

XAI research commonly refers to the trust that the model is correct \citep[e.g.,][]{lipton2016mythos,fidelity,schmidt2019trust}. What does this mean, exactly? 

To illustrate this question, consider some binary classification task, and suppose we have a baseline that is completely random by design, and  
a trained model that achieves the performance of the random baseline (i.e., 50\% accuracy in this case).\footnote{Assume that the performance evaluation is representative of real usage for now, although this is an important factor that we will discuss in Section~\ref{subsec:extrinsic-trust}.}
Since the trained model performs poorly, a simple conclusion to draw is that we cannot trust this model to be correct. But is this true?

Suppose now that the trained model with random baseline performance does \textit{not} behave randomly.
Instead, it is biased in a specific manner, and this bias can be revealed with an interpretation or explanation of the model behavior.
% \yg{I'd add a \emph{concrete} example here, e.g. something like: ``as a cartoon example, consider a credit scoring model that is correct with high probability for loans below some amount, or for certain identified sub-populations.''} 
This explanation reveals to the user that on some types of samples, the model---\textit{which maintains random baseline performance}---is more likely to be correct than for others.\footnote{E.g., calibrated probabilities \citep{lichtenstein1977calibration}, where the classification probabilities of a model are calibrated with some measure of its uncertainty, can produce this effect.} As an illustrative example, consider a credit-scoring AI model that is more likely to be correct for certain sub-populations. 

The performance of the second model did not change, yet we can say that now, with the added explanation, a trustor may have more trust that the model is correct (on specific instances). What has changed? The addition of the explanation enabled the model to be more \textit{predictable}, such that the user can now better anticipate whether the model's decision is correct or not for given inputs (e.g. by looking at whether the individual is part of a certain sub-population), compared to the model without any explanation. Note that this is merely refers to one `instance' of anticipation; it refers to anticipating a particular attribute of the AI's decision (correctness), whereas in the previous definition (\S\ref{sec:trust-basic-definition}), it refers to general behavior. 

% \tm{I think here there is an ambiguity here that needs to be clarified. At the end of Section 2 (final paragraph), the bit about predictability hints that the goal is for the trustor to predict the behaviour of the AI model; but in this section (paragraph above), predictability refers to the ability to predict whether decision is correct. It's not clear to me whether this difference is deliberate on your behalf as it does build the narrative quite well, but if so, I'm not convinced the reader will note the difference without it being explicitly pointed out. Worst case is that this misunderstand this entire section} \todo{I think I agree, I added a few sentences above to clarify this}

We arrive at a more nuanced and accurate view of what ``trust in model correctness'' refers to: it is in fact not trust in the general performance ability of the model, but that \textit{the patterns that distinguish the model's correct and incorrect cases are available to the user}.

% \tm{I have added the paragraph below on distrust }
% The ability to identify incorrect cases is an important part of building trust, or more accurately, \emph{distrust}. This allows the user to identify when a model output should be trusted, and when it should not be. We adapt \citeauthor{tallant2017contractual}'s definition of distrust: A distrusts B if A does not accept vulnerability to B's action because A believes that B may not act in A's best interest \cite{tallant2017contractual}. For the remainder of this paper, we focus our analysis on trust as the link to distrust is straightforward. \todo{I really like this, I'm considering moving this paragraph to sec9 though since I think it hurts the pacing the pacing of this section here}

\begin{table*}
\caption{The European requirements for trustworthy AI, related available documentation, and related explanatory methods or analyses. We position the European guidelines as a guidance on which \textit{contracts} (\S\ref{sec:contractual-trust}) it is useful to pursue trust. 
% \ana{Please fill in the missing values if you know of an appropriate documentation/method.}
}
\resizebox{\textwidth}{!}{
%\begin{tabular}{p{2.75cm}lp{5cm}l}
\begin{tabular}{p{0.15\textwidth}p{0.33\textwidth}p{0.2\textwidth}p{0.375\textwidth}}
\toprule
\multicolumn{2}{c}{\textbf{European Guidelines for Trustworthy AI Models}} & \multirow{2}{*}{\textbf{Documentations}} &\multirow{2}{*}{\textbf{Explanatory Methods/Analyses}} \\
\cmidrule{1-2} 
\textit{Key Requirements} & \textit{Factors} &  & \\
\midrule
\multirow{3}{2.75cm}{Human agency and oversight} & $\cdot$ Foster fundamental human rights & Fairness checklists  & See ``Diversity, non-discrimination, fairness''\\
& $\cdot$  Support users' agency & All & User-centered explanations \cite{Ribera2019CanWD} \\
& $\cdot$  Enable human oversight & N/A & Explanations in recommender systems \cite{kunel_let_me_explain_2019} \\
\midrule
\multirow{5}{2.75cm}{Technical robustness and safety} & $\cdot$ Resilience to attack and security & Factsheets (security)  & Adversarial attacks and defenses \cite{DBLP:journals/corr/GoodfellowSS14} \\
& $\cdot$ Fallback plan and general safety & N/A & N/A \\
& $\cdot$ A high level of accuracy & Model cards (metrics)& N/A \\
& $\cdot$ Reliability & Factsheets (concept drift) & Contrast sets \cite{gardner2020-contrast-sets}, behavioral testing \cite{ribeiro-etal-2020-checklist}\\
& $\cdot$ Reproducibility & Reproducibility checklists & ``Show your work'' \cite{dodge-etal-2019-show} \\
\midrule
\multirow{3}{2.75cm}{Privacy and data governance}  & $\cdot$ Ensure privacy and data protection & Datasheets/statements & Removal of protected attributes \cite{ravfogel-etal-2020-null} \\
& $\cdot$ Ensure quality and integrity of data & Datasheets/statements & Detecting data artifacts \cite{gururangan-etal-2018-annotation}\\
& $\cdot$ Establish data access protocols &  Datasheets/statements & N/A \\ 
\midrule
\multirow{7}{2.75cm}{Transparency} & $\cdot$ High-standard documentation & All & N/A \\
& $\cdot$ Technical explainability & Factsheets (explainability) & Saliency maps \cite{Simonyan2014DeepIC}, self-attention patterns \cite{kovaleva-etal-2019-revealing}, influence functions \cite{pmlr-v70-koh17a}, probing \cite{ettinger-etal-2016-probing}\\
& $\cdot$ Adaptable user-centered explainability & Factsheets (explainability) & Counterfactual  \cite{pmlr-v97-goyal19a}, contrastive \cite{miller2018contrastive}, free-text \cite{Hendricks2018GroundingVE, marasovic-etal-2020-natural}, by-example \cite{pmlr-v70-koh17a}, concept-level \cite{ghorbani2019concept-explanations}  explanations \\
& $\cdot$ Make AI systems identifiable as non-human & N/A  & N/A \\
\midrule
\multirow{3}{2.75cm}{Diversity, non-discrimination, fairness} & $\cdot$ Avoid unfair bias & Fairness checklists & Debiasing using data manipulation \cite{sun-etal-2019-genderbias}\\
& $\cdot$ Encourage accessibility and universal design & N/A & N/A \\
& $\cdot$ Solicit regular feedback from stakeholders & Fairness checklists & N/A \\
\midrule
\multirow{4}{2.75cm}{Societal and environmental well-being} & $\cdot$ Encourage sustainable and eco-friendly AI & Reproducibility checklists & Analayzing individual neurons \cite{Dalvi2019WhatIO}\\
& $\cdot$ Assess the impact on individuals & Fairness checklists & Bias exposure \cite{stanovsky-etal-2019-evaluating} \\
& $\cdot$ Assess the impact on society and democracy  & Fairness checklists & Explanations designed for applications such as fact checking \cite{atanasova-etal-2020-generating} or fake news detection \cite{lu-li-2020-gcan}  \\
\midrule
\multirow{5}{2.75cm}{Accountability}  & $\cdot$ Auditability of algorithms/data/design  & Factsheets (lineage) & N/A \\
& $\cdot$ Minimize and report negative impacts  & Fairness checklists & N/A \\
& $\cdot$ Acknowledge and evaluate trade-offs & N/A & Reporting the robustness-accuracy  trade-off  \cite{melis_alvarez_2018} or the simplicity-equity trade-off \cite{Kleinberg2019SimplicityCI}  \\
& $\cdot$ Ensure redress & Fairness checklists & N/A \\
\bottomrule
\end{tabular}
}

\label{tab:contracts}
\end{table*}

\subsection{The General Case: Trust in a Contract}

The above example of model correctness is merely an instance of what  \citet{hawley2014trust-contractual} and \citet{tallant2017contractual} refer to as \textit{trust with commitment} or \textit{contractual trust}. Contractual trust is when  a trustor has a belief that the trustee will stick to a specific contract.%Contractual trust is a model of trust as a triplet of the trustor, the trustee, and a \textit{contract}---as a belief that a specific commitment will be fulfilled, i.e. ``I trust that you will do X.''
\footnote{To our knowledge \citet{hawley2014trust-contractual} is the first to formalize  trust as ``trust with commitment [= contract].'' \citet{tallant2017contractual} expands on their work with terminology of contractual trust.}

In this work, we contend that all \HumanAI trust is contractual,\footnote{Although they do not refer to contractual trust in their work, \citet{hoffman2017taxonomy-trust-automation} provide support to formalize trust in automation (beyond AI) %human-machine \ana{\HumanAI ?} \todo{the cited work is about trust in machines in general, any automation} trust
as multi-dimensional (which we interpret as multi-contractual), rather than a binary variable or sliding scale.} 
and that regardless of what the contract is in a particular interaction, to discuss \HumanAI trust, \textbf{the contract must be explicit}. 

Generally, the contract may refer to any functionality that is deemed useful, even if it is not concrete performance at the end-task that the model was trained for. Therefore, model correctness is only one instance of contractual trust. For example, a model trained to classify medical samples into classes can reveal strong correlations between attributes for one of those classes, giving leads to research on causation between them, even if the model was not useful for the original classification task \citep{lazer2014parable,lipton2016mythos}. %\ana{Changed after ``even if...''. Please check.} \todo{i don't agree with "due to poor performance" because in some cases, people specifically train models on a related use-case just to derive correlations/causations, and performance is not really important} %.even if the model was not directly useful in the classification use-case \citep{lazer2014parable,lipton2016mythos}.
% \tm{Great example}

% \tm{I have added the following paragraph, but feel free to remove if you think it breaks the flow. It does seem important to address though}

\noindentparagraph{\textbf{Contracts and contexts.}} The idea of context is important in trust: people can trust something in one context but not another \cite{hoffman2017taxonomy-trust-automation}. For example, a model trained to classify medical samples into classes can perform strongly for samples that are similar to those in its training set, but poorly on those where some features were infrequent, even though the `contract' appears the same. Therefore, contractual trust can be stated as being conditioned on context. For readability in the rest of this paper, we omit context from the discussion, but implicitly, we consider the contract to be conditioned on, and thus include, the context of the interaction.

% Following this, we complete the  translation of the stated goal with a final revision:

% \noindentparagraph{Claim D (final).} A goal of explainability/interpretability is to increase the trustworthiness of the AI model so that the user develops (warranted) trust in \textit{the stated contract}.

\noindentparagraph{\textbf{What are useful contracts?}}
%The AI HLEG group appointed by
The European Commission %(the executive branch of the EU)
has outlined detailed guidelines on what should be required from AI models for them to be trustworthy (see Table \ref{tab:contracts}, col.\ 1--2).\footnote{The guidelines are available at \url{https://ec.europa.eu/digital-single-market/en/news/ethics-guidelines-trustworthy-ai}.} Each of these requirements can be used to specify a useful contract.

Another area of research that is relevant for defining contracts is the work that proposes standardized documentations to communicate the performance characteristics of trained AI models. The examples of such documentations are: data statements \citep{bender2018data-statements}, datasheets for datasets \citep{Gebru2018DatasheetsFD}, model cards \citep{mitchell2019model-cards}, reproducibility checklists \citep{pineau_reproducibility}, fairness checklists \citep{Madaio2020CoDesigningCT}, and factsheets \citep{arnold2019factsheets}. 

We illustrate the connection between these documentations and the European requirements in Table \ref{tab:contracts}. For example, if transparency is the stated contract then all of the mentioned documentations could be used to specify information that AI developers need to provide such that they can evaluate and increase users' trust in transparency of an AI system. 

\noindentparagraph{\textbf{Explanation and analysis types depend on the contract.}} We argue that ``broad trust'' is built on many contracts, each involving many factors and requiring different evaluation methods. For example, the models' efficiency in terms of the number of individual neurons responsible for a prediction is relevant for sustainability, but likely not for, e.g., ensuring universal design. %accessibility. 

We have previously illustrated that the addition of explanation of the model's behavior can increase users' trust based on one contract (\sect{sec:trust_model_correctness}). Just as different evaluation methods are needed for different types of contractual trust, so are different types of explanations. In Table \ref{tab:contracts}, we outline different established types of explanatory methods and analyses that could be suitable for increasing different types of contractual trust derived from the European requirements.

\noindentparagraph{\textbf{Conclusions.}} The formalization of contracts allows us to clarify the goal of anticipation in \HumanAI trust: contracts specify the behavior to be anticipated, and \textbf{to trust the AI is to believe that a set of contracts will be upheld}. 
% \tm{I emphasised the previous sentence -- it seems important enough to do so, but feel free to remove}

%\noindentparagraph{Contracts Taxonomy.}
Specific contracts have been outlined and explored in the past when discussing the integration of AI models in society. We advocate for adoption of the taxonomy of contracts in \HumanAI trust, for three reasons: (1) it has, though recent, precedence in sociology; (2) it opens a general view of trust as a multi-dimensional transaction, for which all relevant dimensions should be explored before integration in society; and importantly, (3) the term implies an \textit{obligation} by the AI developer to carry out a prior or expected agreement, even in the case of a social contract.

\section{Trustworthy AI}
\label{sec:trust-trustworthy}

% We have established predictability as a goal of \HumanAI trust as the belief that an explicitly stated contract will be upheld.\ana{This is hard to parse.} 
Trust, as mentioned, aims to enable the ability to anticipate intended behavior through the belief that a contract will be upheld. 
Further, as mentioned in Section~\ref{sec:trust-basic-definition}, the ability to anticipate does not necessarily manifest with the existence of trust; it is possible for a user to trust a model despite their inability to anticipate its behavior. In other words, the belief exists, but may or may not be followed by the desired behavior. What differentiates trust that `succeeds' at this goal from trust that does not?

Let us separate the two cases from the perspective of AI: given that the user trusts the AI model, anticipation depends on \textit{whether the model is able to carry out its contract.} This perspective distinguishes ``trust'' (an attitude of the trustor) from being ``trustworthy'' (a property of the trustee) \citep{Solomon1998-SOLCTI,Wright2010-WRITAT,thiebes2020trustworthy}, and we say that \textbf{an AI model is trustworthy to some contract if it is capable of maintaining this contract}. 

Trust and trustworthiness are entirely disentangled: pursuing one does not entail pursuing the other, and trustworthiness is \textit{not} a prerequisite for trust, in that trust can exist in a model that is not trustworthy, and a trustworthy model does not necessarily gain trust. 
% We categorize trust as \textit{warranted} or \textit{unwarranted} if it is caused by trustworthiness or not \citep{stanford-trust}.
We say that the trust is \textit{warranted} if it is the result of trustworthiness, and otherwise it is \textit{unwarranted} \citep{stanford-trust}.
Warranted trust is sometimes referred to as trust that is \textit{calibrated} with trustworthiness \citep{lee2004trust-calibrated}. In other words, trust is the cognitive mechanism to give the `illusion' of anticipating intended behavior, which becomes reality when the trust is warranted, and the trustor feels ``betrayed'' when the illusion is broken.

%\ana{Proposal: make two diagrams with clearly stated cause and effect, one diagram for interpersonal trust, and the other for \HumanAI trust.} \todo{i made one for human-ai (always planned on doing it) but what do you mean by diagram for interpersonal trust?}\ana{The interpersonal trust was before described as an effect, but now that doesn't seem to be important.}

\begin{figure}[t]
\centering
\includegraphics[width=0.85\linewidth]{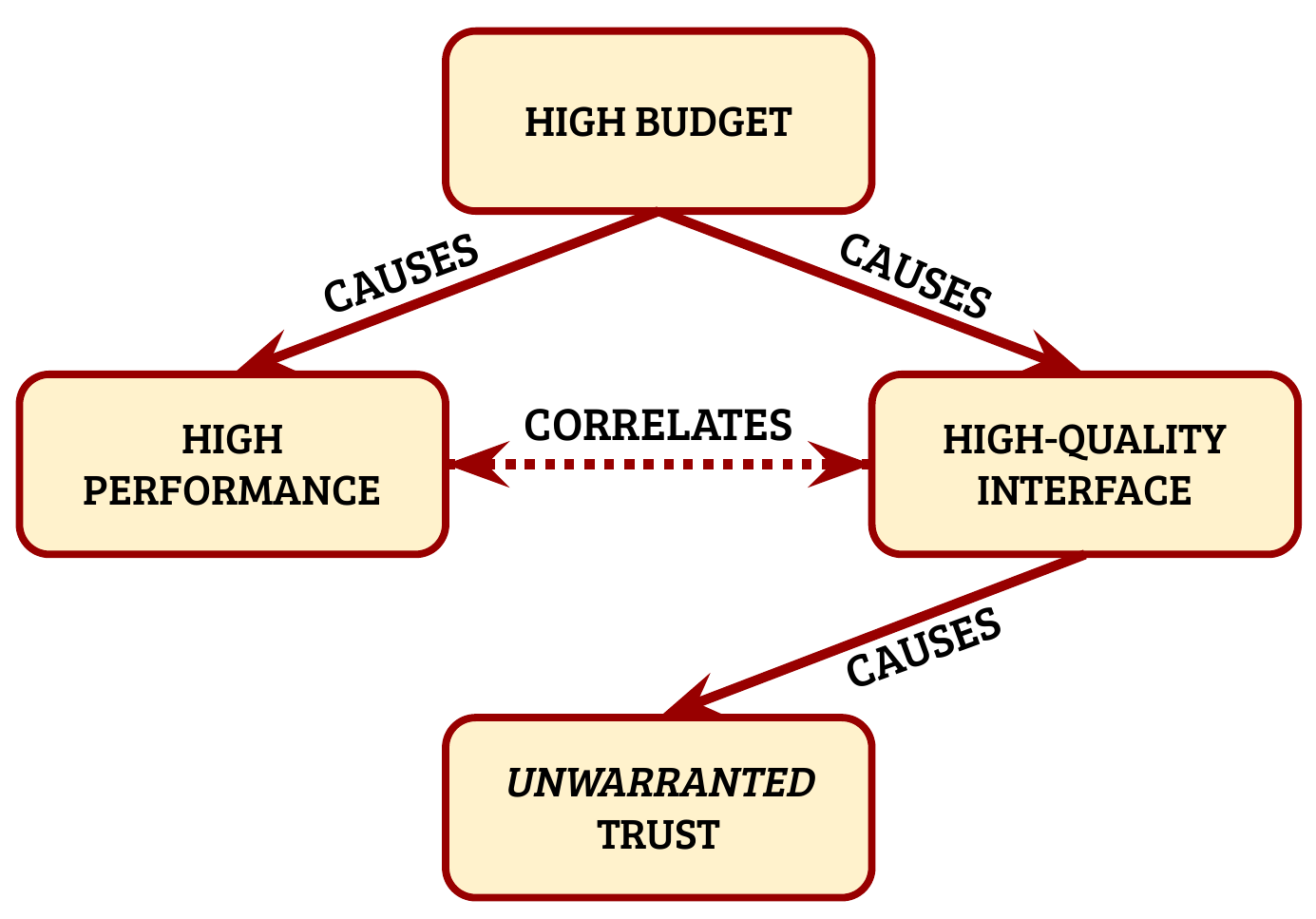}
\caption{An example of causes of trust, in the context of warranted and unwarranted trust. For a contract of model performance, \textit{high-quality interface} is not a cause of high performance, and therefore, any trust that the user gains as a result of the interface is unwarranted.}

\label{fig:unwarranted-ex}
\end{figure}

For example, consider a user interacting with an AI model via some visual interface (GUI), and the user trusts the AI model to make a correct prediction on some task. There is a correlative, but not causal, relationship between high-quality GUI and trustworthy AI models due to a shared variable of high budget in the system development. If the cause of the user's trust is the model GUI, then manipulation of the model's ability to make good predictions will not affect this trust, and thus it is \textit{unwarranted}. If the cause of the trust is the model's performance ability (due to its higher budget), then theoretically manipulating this performance level will affect the level of trust (Figure~\ref{fig:unwarranted-ex}). For instance, \citet{DBLP:journals/corr/abs-1810-05798} show a case where the interface can increase doctors' confidence in a tool, despite not significantly increasing their accuracy.

Formally, we define warranted \HumanAI trust via a causal (interventionist) relationship with trustworthiness: incurred \HumanAI trust is warranted if the trustworthiness of the model can be theoretically manipulated to affect the incurred trust. Note that by this definition, \textit{it is possible for a trustworthy model to incur unwarranted \HumanAI trust}---in this case, the trust will not be betrayed, even though it is unwarranted. %\ana{This example is helpful, consider providing it sooner to make definitions easier to follow.} \todo{where did you mean by earlier?} \ana{I moved this paragraph. It was before the GUI example previously.} 

\begin{figure*}[t]
\centering
\includegraphics[width=0.65\textwidth]{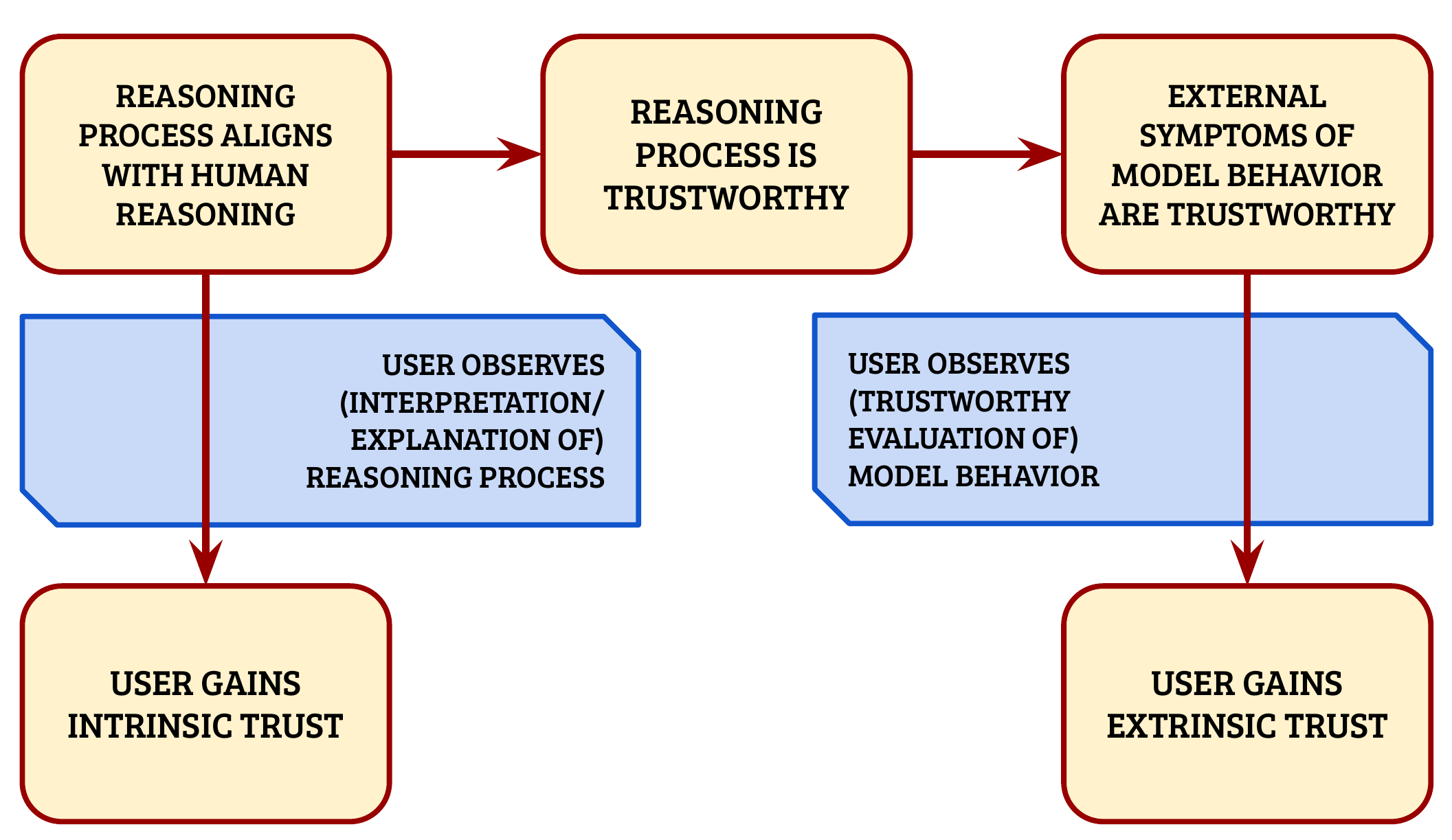}
\caption{A schematic of the causes of \textit{warranted} trust in AI models (summary of \S\ref{sec:causes-of-trust}). 
% \todo{change "reasoning process resembles human reasoning" to "reasoning process is correct" or "agreeable" maybe?}
}
\label{fig:causes}
\end{figure*}

When XAI literature refers to trust, we assume that it is referring to trust that is warranted.\footnote{Not all researchers and organizations may be interested in the distinction between warranted and unwarranted trust, unless required by regulation. We contend that this is an ethical issue. In this work, we assume that it is a core motivation of the field to remain ethical and not implement or prioritize unwarranted trust.} 
Therefore, we contend that \textbf{when pursuing \HumanAI trust, unwarranted trust should be explicitly evaluated against, and avoided or otherwise minimized.} See \cite{parasuraman1997automation-use-misuse} for analysis on the dangers of human usage (or avoidance) of automation when trust and trustworthiness are misaligned, on axes of disuse, misuse and abuse. Specifically, trust exceeding trustworthiness leads to misuse, while trustworthiness exceeding trust leads to disuse.

Finally, the notion of warranted distrust is similar to that of warranted trust, and easily derived from it: we say that the distrust is warranted if it is sourced in the non-trustworthiness of the AI, i.e., the lack of capability of the AI to maintain the contract, and otherwise, it is unwarranted. It stands to ethics that \textbf{if an AI model is incapable of maintaining some relevant contract, it is a desired outcome (desired by the developer) that the user develop warranted distrust in that contract}, that will be beneficial in applying the AI model to some scenario despite its flaws.

\section{Defining Human-AI Trust}
\label{sec:human_ai_definitions}
This section serves as a formal definition of the taxonomy thus far.
Note that we consider \textbf{\textit{AI}} as an automation
that is attributed with human-like intelligence by the human interacting with it.\footnote{We consider AI as an automation attributed with intent, and thus, \textsc{Human-AI} trust is a sub-case of \textit{human-machine} trust \cite{parasuraman1997automation-use-misuse,hoffman2017taxonomy-trust-automation}. Trust in automation that is not anthropomorphized can be considered as \textit{reliance} (\S\ref{subsec:reliance}).}

%\ana{Changed ``machine'' to ``tool'' to avoid confusion with trust in automation.} \todo{AI is a subcase of automation though, so why would it be confusing?} \ana{Because of the work referenced in Section 9.4. I was afraid someone might think we are re-inviting the wheel then. No strong opinion though. :) } \todo{tried adding a footnote here, does that help?} \ana{Looks good.}

%\noindentparagraph{Trustworthy AI.} 
\noindentparagraph{\textbf{Trustworthy AI.}} An AI model  is trustworthy to contract C if it is capable of maintaining the contract.

%\noindentparagraph{\HumanAI trust.} 
\noindentparagraph{\textbf{\HumanAI trust.}} If H \textit{(human)} perceives that M \textit{(AI model)} is trustworthy to contract C, and accepts vulnerability to M’s actions, then H trusts M contractually to C. The objective of H in trusting M is to anticipate that M will maintain C in the presence of uncertainty, and consequently, \textit{trust does not exist} if H does not perceive risk.

Previously, we note that a user's ``anticipation'', defined as the user's belief that the AI will work `as intended', is a key aspect of interpersonal trust that we will use to define \HumanAI trust. Indeed, ``anticipation'' occurs when a user believes that an AI model is capable of maintaining the contract, i.e., a user believes that the model is trustworthy to the contract, which is a prerequisite of \HumanAI trust.

%\noindentparagraph{Warranted and unwarranted \HumanAI trust.}
\noindentparagraph{\textbf{Warranted and unwarranted \HumanAI trust.}} H's trust in M (to C) is warranted if it is \textit{caused} by trustworthiness in M. This holds if it is theoretically possible to manipulate M's capability to maintain C, such that H's trust in M will change. Otherwise, H's trust in M is unwarranted. 

\noindentparagraph{\textbf{\HumanAI distrust.}}  If H \textit{(human)} perceives that M \textit{(AI)} is not trustworthy to contract C, and therefore does not accept vulnerability to M’s actions, then H distrusts M contractually to C. We say that it is \textit{warranted} distrust if the distrust is caused by the non-trustworthiness of M.

%\ana{I kind of want to put these three definitions in a special box colored differently (light grey). Is that too much? :D} \todo{that sounds fun but I dunno how to do it! maybe when this paper is converted into a book...}

\section{Causes of Trust} \label{sec:causes-of-trust}

The next natural question to ask is on the cause behind trust in an AI model. As established earlier, we say that trustworthiness is a prerequisite to warranted trust. What causes a model to be trustworthy? And what enables trustworthy models to incur trust?

We divide causes of warranted trust into two types: intrinsic and extrinsic.\footnote{The distinction, and choice of terminology, are ours. Note that this is unrelated to `intrinsic or enforceable' trust as coined by \citet{gertjan2006intrinsic-and-enforceable-trust}.}

\subsection{Intrinsic Trust} \label{subsec:intrinsic-trust}

A model is more trustworthy when the observable decision process of the model matches user priors on what this process should be.
% \footnote{Faithful causal attribution of behavior aligned with social attribution of intent \citep{miller2017social,jacovi2020aligned-social}.}
% \ana{``faithful causal attribution of behavior aligned with social attribution'' is not clear if you're not familiar with the provided references.} 
This is equivalent to, for example, a doctor that is considered more trustworthy because they are citing various respectable studies to justify their claims.

% The field of explainability in AI aims to explain the decision process of the AI to the user, such that they can comprehend it as they do the decisions of people. The process of explaining involves two steps of \textit{causal attribution} and \textit{social attribution} \cite{miller2017social}, in which the user comprehends the causal chain of events behind the final decision, and attributes social intent to the causal chain. 

% Causal and social attributions are a prerequisite, and not a cause, behind intrinsic trust. 
% Only after these attributions match the user prior, the trust is gained. 

Explanation in AI aims to explain the decision process of the AI to the user, such that they can understand why and how the decision was made. However, the process of explaining does not, in itself, enable intrinsic trust. Only when (1) the user successfully comprehends the true reasoning process of the model, and (2) the reasoning process of the model matches the user's priors of agreeable reasoning, intrinsic trust is gained.

% For example, a decision tree is a model whose inner workings can be well-understood by the user (if it is sufficiently small). \textbf{If the user has no prior on what behavior is trustworthy for the given task, intrinsic trust will not be gained, even if the AI is clear and easy to understand}: for example, for a task involving complex expert knowledge, a layman user will not be able to gain intrinsic trust in the model regardless of how `simple' and interpretable the model is.  
% \yg{I'd switch the order of the bold part and the part that follows, with something like ``however, for a task involving complex ... . That is, \textbf{if the user has no prior on ...}'' }\ana{+1}

For example, a decision tree is a model whose inner workings can be well-understood by the user (if it is sufficiently small). However, e.g., for a task involving complex expert knowledge, a layman user will not be able to gain intrinsic trust in the model regardless of how `simple' and interpretable the model is. \textbf{If the user has no prior on what behavior is trustworthy for the given task, intrinsic trust will not be gained, even if the AI is easy to understand.}

Further of note is the granularity of interpretability that will facilitate intrinsic trust. The issue of enumerating relevant priors to the user through explanation is not trivial: it is possible to convey information about the reasoning process that is accurate to the model and easy to comprehend, but nevertheless irrelevant to the priors that the user cares to know about. In the same way, it may be possible to derive an explanation of the reasoning process at very coarse granularity and remain descriptive of the relevant priors.

% Of course, such priors do not have to be complete or particularly well-developed to be useful for intrinsic trust. 
In the example of the doctor citing respectable studies, it is not necessary for the user to be able to understand those studies---merely that they are respectable, and that their claims match the claims of the doctor. As another example, ``the model should not make any decision 
% \yg{about what? for exmaple, for authorship attribution, or for mental health prediction, or for ``content vs boilerplate'' etc, the number of commas can be predictive. for sentiment, the number of exclamation marks can be predictive, etc.} 
based on the number of commas in the text'' can be such a prior, if the user truly perceives ``the number of commas'' as irrelevant to the decision. \textbf{Generally, we regard any prior as relevant so long as it is useful to upholding the contract in question.} For example, the absence of heuristical behavior (such as detecting the number of commas) can be such a prior towards a contract of model correctness.
% , such as the negation of suspicious behavior as a prior towards distinguishing model correctness. 

Intrinsic trust can be increased in a disciplined manner by formalizing the priors behind trustworthy or non-suspicious behavior, and incorporating behavior that upholds those priors. The documentations outlined in Table \ref{tab:contracts} (\sect{sec:contractual-trust}) could be useful for that.

% \todo{counterfactual/by-example explanations i would also count under intrinsic, maybe worth mentioning here}

%\ana{Having priors vs.\ human agency?}

\subsection{Extrinsic Trust} \label{subsec:extrinsic-trust}

It is additionally possible for a model to become trustworthy not through explanation, but through behavior: in this case, the source of trust is not the decision process of the model, but \textit{the evaluation methodology} or \textit{the evaluation data}. This is equivalent to a doctor who is considered more trustworthy because they have a long history of making correct diagnoses; or because they graduated from a prestigious institute that is considered to have rigorous student evaluation. The trust comes from observing symptoms of a trustworthy model, rather than the model's inner working. 
% It is similar in principle to behavioral/black-box testing of automation \citep{beizer1995blackboxtesting}.

Extrinsic trust can be seen as the inverse of intrinsic trust: where aligned priors cause a trustworthy model, a trustworthy model causes convincing external symptoms. Since both share a causal relationship with a trustworthy model, they are both used by people as indicators of trustworthiness and thus incur warranted trust. This perspective is outlined in Figure~\ref{fig:causes}.

For AI models, \textbf{extrinsic trust is, in essence, trust in the evaluation scheme}. To increase extrinsic trust is a matter of justifying that a model can generalize to unseen instances based on expected behavior of the model on other unseen instances. 

Extrinsic trust is gained by two independent requirements: (1) when the model is trustworthy, and (2) \textit{when the evaluation scheme is trustworthy}. To show that some evaluation scheme is `trustworthy' is to justify that the distribution of instances during evaluation matches the distribution of the true unseen instances that will require trust (in a specific contract) by the user in the future---or in other words, guarantee that it is only possible for the AI to pass the evaluation if it is capable of maintaining the contract.\footnote{E.g., an evaluation scheme that verifies whether the AI does not discriminate against a sub-population may be different from an evaluation scheme that verifies general performance ability.}
% \yg{this may be obvious by ``in a specific contract'', but I'd still like to have a concrete example showing two different kinds of trust, ie, for the model to be accurate above some accuracy, vs. eg for the model to not discriminate against some subpopulation. This appears later, but I think should appear already here, even if just briefly.}

Three main methods of evaluation towards extrinsic trust:

%\noindentparagraph{By proxy.} 
\noindentparagraph{\textbf{(1) By proxy.}} Expert (human) opinion on AI reasoning or behavior can enable non-experts to gain extrinsic trust in the AI. Note that the expert does not necessarily gain trust at this point, because the expert may or may not be vulnerable to the AI's decisions, as the interaction between the expert and the AI is made under different terms than that between the AI and the user. Also of a note is that what exactly constitutes a trustworthy expert for this purpose is a question of interpersonal trust, and not \HumanAI trust.
%\footnote{\ana{Since we introduced ``trust in contract'' I feel like it would be useful to spell out why are the experts not vulnerable to the AI's decisions. Developers' deploy models that they have behaviourally tested knowing that users might be disappointed by unforeseen bugs, but they take that risk. In this case, the contracts are the different types of tests the developer used to behaviourally test the models, and in an ideal world the developer deploys the model only when the developer is ready to accept the consequences of deploying (i.e., vulnerability). But what if the developer is rushed to meet a deadline (realistic case)? In this case, despite the model being contractually trustworthy, it's not necessarily that the developer trusts the model. Is this too tangential?} \todo{changed the text to address this, i hope?}}

%\noindentparagraph{Post-deployment data.}
\noindentparagraph{\textbf{(2) Post-deployment data.}} Most simply, the examples that the model sees during production are the most trustworthy representatives of general behavior evaluation, notwithstanding issues of distribution shift over time. Such examples may or may not have gold supervision, and performance may be measured by some weaker signal. Although the distribution of these examples is realistic, it is not controllable, and thus there is little control on the specification of contracts to be evaluated. 

%\noindentparagraph{Test sets.}
\noindentparagraph{\textbf{(3) Test sets.}}Sets of examples, distributed in some specific way, for which gold labels are available. Test sets are generally seen as imperfect approximators of post-deployment data, as their distribution is not representative of true unseen data \cite{DBLP:journals/corr/abs-2007-14435} 
% \ana{Perhaps: ``Towards Ecologically Valid Research on Language User Interfaces''? or ``Evaluating Machines by their Real-World Language Use" a good example for "post-deployment data''?} 
and thus cannot imply overall generalization.

However, this perspective only takes into account one type of contract: high, general, post-deployment performance. The distribution of the test examples is controllable, such that \textit{they can specialize in specific contracts}, such as specific phenomena, robustness and adversarial robustness, fairness and ethics, privacy, and so on. For example, by showing that a model performs equally for two large (and convincingly distributed) collections of people of different races, a user may deem the model more trustworthy to the contract: ``I trust you to not discriminate based on race.''
Previous work has used specialized test sets to verify particular contracts \cite{DBLP:conf/starsem/KiritchenkoM18,barnes-etal-2019-sentiment-not-solved,olmpics}, or outlined methodologies for constructing such evaluation \citep{kaushik2020-counterfactual,gardner2020-contrast-sets,ribeiro-etal-2020-checklist,anders2020talk-random-splits}. 

% \todo{how can we better formalize how good is a particular test set is at satisfying a particular contract?} \ana{This is important.} \todo{need to look for the descriptions of social constructs vs mathematical definitions to measure them}

% \tm{Perhaps we could formalise by noting how well is holds to the assumptions underlying the use of models or something similar? Using test sets to validate anything assumes two things: (1) that the test data accurately represents the underlying mechanism it comes from (e.g. the labels need to be `gold', the distribution needs to be complete for the contract, etc.); and (2) that this underlying mechanism will be the same in the future as it was in the past (when the data was derived). The degree to which these assumptions hold affects the trustworthiness of the contract. Or maybe I'm missing what you mean here?}

% [tim 1. 2. + 3. mathematical (cite lipton, suresh)]

\noindentparagraph{\textbf{How can we verify whether an evaluation scheme (in particular, test sets and deployment data) is trustworthy?}} Using data for validation assumes the following: (1) that the data accurately represents the underlying mechanism it comes from (e.g., the label is correct, the distribution is representative for the contract at the time of data collection); (2) that the underlying mechanism is negligibly affected by distribution shift over time; and (3) that the \textit{evaluation metrics} represent the contract---i.e., that a positive result implies the capability to maintain the contract, and the inverse. The degree to which these assumptions hold affects the validity of the evaluation scheme.

Notably, point (3) is affected by the `accurate' formalism of contracts. For example, there are multiple formal measures of fairness such as individual fairness, demographic parity and equalized error rates \cite{10.1145/3294052.3322192}. However we cannot say that each one of these measures completely encapsulates the social requirement of fairness: each measure formalizes a different aspect of fairness, and there cannot be a solution that satisfies all of them.

\section{Explainability and Trust} \label{sec:explainability-trust}

As mentioned, the following is a common claim of XAI literature:

% \yg{I find it confusing that claim A and B have the same title. I think we should distinguish them. Or not stating the first as an official ``claim''. Also, is it a claim or a definition?} \todo{i tried to give them titles below, but I will definitely accept better suggestions}

\vspace{0.2cm}
\noindent\textbf{XAI for Trust \textit{(common)}:} A key motivation of XAI and interpretability is to increase the trust of users in the AI. 
\vspace{0.2cm}

However, the above claim is difficult to probe without a definition of trust. Following the formalization in this work, we can now elucidate this claim with details on the true nature of this motivation:

\begin{figure*}[t]
\centering
\includegraphics[width=0.9\textwidth]{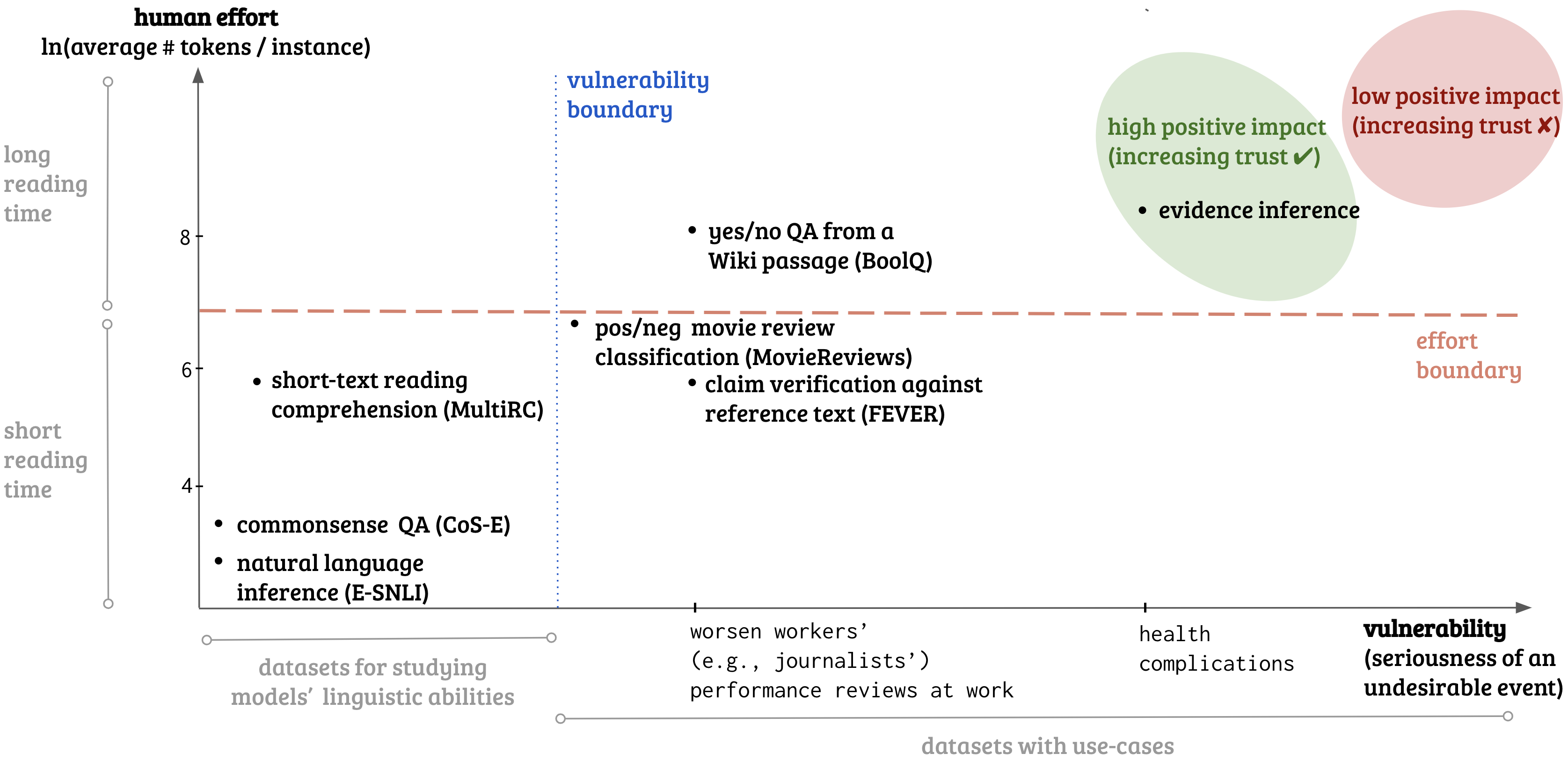}
\caption{Categorization of datasets that are commonly used to advance interpretable NLP with respect to required effort from people to complete associated tasks (y-axis) and seriousness of an undesirable event (x-axis). %The number next to each dataset reports the average number of tokens per instance, which we take to be a proxy of the task difficulty for humans. 
Datasets that require long reading time and that have associated use-cases are better suited for studying trust. The average number of tokens per instance are: e-SNLI (16), CoS-E (28), MultiRC (303), FEVER (327), MovieReviews (774), BoolQ (3583), Evidence Inference (4761). 
}
\label{fig:eraser_categorization}
\end{figure*}

\vspace{0.2cm}
\noindent\textbf{XAI for Trust \textit{(extended)}:} A key motivation of XAI and interpretability is to (1) increase the trustworthiness of the AI, (2) increase the trust of the user in a trustworthy AI, or (3) increase the distrust of the user in a non-trustworthy AI, all corresponding to a stated contract, so that the user develops warranted trust or distrust in that contract.

Let us clarify this claim by unraveling it:

% \pagebreak

\vspace{0.2cm}
\noindent\textbf{\textit{A key motivation of XAI and interpretability is ...}}

\vspace{0.2cm}
\noindent\textbf{\textit{... to (1) increase the trustworthiness of the AI ...}}

AI is said to be trustworthy to a contract if it is capable of maintaining the contract. Then XAI is a method of \textit{creating} a capability by revealing the relevant signals in the AI reasoning process (as in the example of the random baseline-like model, \S\ref{sec:trust_model_correctness}). An AI model that hides these signals would be less trustworthy as they fail to uphold some contracts (e.g., Table \ref{tab:contracts}).

% \tm{I have added the last sentence above}

\vspace{0.2cm}
\noindent\textbf{\textit{... (2) increase the trust of the user in a trustworthy AI ...}}

The goal of developing trust, from the user's perspective, is to enable the ability to anticipate behavior in the presence of risk. Then XAI is a method of allowing the user easier access to the signals that enable this anticipation.

\vspace{0.2cm}
\noindent\textbf{\textit{... or (3) increase the distrust of the user in a non-trustworthy AI ...}}

Similarly, the user's goal in distrust, in the presence of risk, is to anticipate when desired behavior will \textit{not} happen. This is the inverse of (2), therefore XAI is a method of enabling distrust. 

\vspace{0.2cm}
\noindent\textbf{\textit{... all corresponding to a stated contract ...}}

The above is only relevant with respect to specific contracts that dictate what precisely is anticipated. Therefore, the contract must be explicitly recognized by the XAI methodology, so that it reveals information that is relevant to create or reveal the capability of the AI to maintain the contract.

\vspace{0.2cm}
\noindent\textbf{\textit{... so that the user develops warranted trust or distrust in that contract.}}

For the user to achieve their goal of anticipation, their trust should be warranted, and XAI verifies this by revealing the \textit{intrinsic} or \textit{extrinsic} sources (causes) of the trust.

\section{Evaluating Trust} \label{sec:evaluation}

The last question that remains for us to discuss is on the evaluation of trust. What should evaluation of trust satisfy? Do current methods of evaluating trust fulfill these requirements? 

% Further, in some cases trustworthiness may only be verified based on the pair of model + interpretation, and not simply the model in itself.

\subsection{Vulnerability in Trust and Trustworthiness}

% \yg{Maybe we should be explicit about the claim: ``Evaluation of accuracy of models, as common in the literature and in production settings, may be used to increase external trust in the model for others by asserting trustworthiness or by providing evidence for trustworthiness, but they are \emph{not} on their own evaluating any notion of trust, as there is no risk involved.}
The distinction of evaluation between trust and trustworthiness stems from their distinction on vulnerability:
trustworthiness does not require the trustor to accept risk to manifest. For instance, the evaluation of the accuracy of AI models may be used to increase external trust in the model for others by asserting trustworthiness or by providing evidence for trustworthiness, but they are \emph{not} on their own evaluating any notion of trust, as there is no risk involved.

% \newpage
%\noindentparagraph{Trust:}
\noindentparagraph{\textbf{Vulnerability in Trust.}} By definition, the question of trust does not exist when the user does not assume a risk in trusting the AI. The user must depend on the AI on some level for risk to manifest: for example, when the user is already confident in the answer and does not actually depend on the AI---as is the case for many machine learning datasets today---there is no unfavorable event that directly results from the AI's actions. 
% For example, \citet{feng-shi2019-triviabowl} show an evaluation setting that incorporates vulnerability, as the human explainee will lose the trivia game if their trust is betrayed; they accept the vulnerability of losing when they interact with the artificial model for advice. \ana{This example is already given before.} \todo{in the same context too, so i think there's no need for an example here.} 
Therefore, \textbf{experiments that simply ask the users whether they trust the model for some trivial task evaluate \textit{neither} trust nor trustworthiness.}

We illustrate this point with the ERASER benchmark \cite{deyoung-etal-2020-eraser} as a case-study (Figure \ref{fig:eraser_categorization}). ERASER is a benchmark for natural language processing interpretability, and consists of multiple  datasets. For the purpose of this illustration, we set  simple measures of vulnerability and required effort for humans to solve associated tasks. We also assume that the task domain (what the AI is trained on) is also the use-case (the real interaction setup), e.g., the user must provide the label for a problem instance, and is advised by an AI model trained to do the same. We heuristically measure required human effort using the length of the input (average number of tokens), expecting that the longer the text is, the more time-consuming and laborious is the task to people, and consequently people are more dependent on AI \cite{NIPS2019_8301}. We set the ``effort boundary''---a threshold for tasks requiring little effort---around 774 tokens (a few minutes of reading). Motivated by categorization in \cite{Schlangen2020TargetingTB}, we set the ``vulnerability boundary''---a hypothetical threshold for tasks without an associated undesirable event--- right from natural language inference \cite{bowman-etal-2015-large}, commonsense QA \cite{talmor-etal-2019-commonsenseqa}, and multi-sentence reading comprehension \cite{khashabi-etal-2018-looking}, as these tasks are designed only to test  models' capabilities in capturing particular linguistic phenomena, and, to the best of our knowledge, there is no undesirable event from interacting with them in the base scenario.

The question of trust can be considered meaningful for tasks above the ``effort boundary'' and right from the ``vulnerability boundary'', and notably, only two to three ERASER tasks can be placed in those areas 
(e.g., evidence inference).\footnote{Given a scientific article describing a randomized controlled trials, evidence inference is the task of inferring whether a given intervention is reported to either significantly increase, significantly decrease, or have no significant effect on a specified outcome, as compared to a comparator of interest.} 

We advocate that only use-cases that can be attributed with both considerable required human effort and vulnerability, are used to target, evaluate and discuss trust. Although in our ERASER example we treat the task domains as use-cases, comprehensive discussion in this area must develop the use-case explicitly. The use-case may not necessarily be an exact replica of the AI's task domain, and thus, the question of vulnerability would depend entirely on the use-case; additionally, we note that the measure of difficulty is not solely constrained to how time-consuming the task is, as even `easy' tasks may require trust if the stakes are high.

%\ana{\citet{Lubars2019AskNW} use the following factors to measure the difficult of tasks: (1) social skills, (2) creativity, (3) effort, (4) expertise, (5) (perceived) human ability. TODO: Review datasets and tasks NLP researchers use to develop and categorize current explanatory methods with respect to these factors, and demonstrate that most of them are designed to explain easy tasks that humans do not need help with. Then suggest tasks we should focus on (those we need help with), with the special emphasize on tasks that could make a real impact (e.g. summarization of clinical trials). Finally, think how can we make this categorization more formal.}

%\noindentparagraph{Trustworthiness.}
\noindentparagraph{\textbf{Vulnerability in Trustworthiness.}} Whether a model is intrinsically or extrinsically trustworthy is unrelated to the existence of vulnerability in the user. For example, a domain expert can verify the intrinsic trustworthiness of a model by verifying whether the model performs the expected reasoning steps to arrive at its decision, despite the expert not necessarily assuming any vulnerability in doing this inspection, the same way that a class teacher evaluates their students.
% \footnote{The conclusion of the expert can in turn increase extrinsic trust in a layman user, as proof that the expert's verification of trustworthiness enabled it into trust in the user.}

\subsection{Warranted and Unwarranted Trust}

Of course, it is impossible to differentiate between warranted and unwarranted trust simply by evaluating whether the user trusts the model. In this area, \citet{Kaur2019InterpretingIU} show a synthetic experimental setup to evaluate unwarranted trust, and conclude that even data scientists are susceptible to develop unwarranted trust, despite some mathematical understanding of the models. \citet{smith-renner2020accountabiity} show similar conclusions on unwarranted trust in a different experimental setup.
%\ana{I am missing an alternative approach here. Perhaps we can expand on something mentioned in Section 4: ``the model can be theoretically manipulated to affect the incurred trust''? If we don't have any ideas, we should perhaps say explicitly that future work should design the evaluation of warranted trust?} \todo{I agree. added the paragraph below, what do you think?}
% This area is underdeveloped, and we hope that future work considers advancing the question of evaluating warranted and unwarranted.
However, this area is underdeveloped.

Our discussion of warranted trust in Sections \ref{sec:trust-trustworthy} and \ref{sec:human_ai_definitions} suggests a possible evaluation based on manipulationist causality---i.e., that if the trust is warranted, the level of trust can be manipulated by manipulating the cause of trustworthiness.
This gives rise to the following evaluation protocol: 
\begin{enumerate}
    \item Measure the level of trust in an interaction.
    \item Manipulate the real trustworthiness of the model (e.g., by handicapping it in some way; by improving its predictions; or even by replacing the model with an oracle).
    \item Measure the level of trust after the manipulation.
\end{enumerate}
The amount of change due to the manipulation indicates the level of warranted trust.

\subsection{Evaluating `Anticipation'}

While not an evaluation of trust, there are methods of evaluating the ability of users to successfully anticipate the AI's behavior, an important aspect of \HumanAI trust, via \textit{simulatability} \citep{doshi2017rigorous-humangrounded,hase2020simulatability}---the ability of the user to simulate the outcome of the AI on an instance level. 
% As previously mentioned, the existence of vulnerability and predictability are indicative of warranted trust. 
Thus, simulatability can serve as one proxy signal to assess whether the goal of trust has been achieved, though we note that it does not concretely verify the existence of trust---on the contrary, it relies on a prior assumption that trust exists.
% . However, note that \textit{for the simulatability results to be useful signal on the existence of warranted trust, risk must be accounted for during the predictability measurement}.
% \yg{what does the last sentence mean, concretely?}
Additionally, since simulatability is performed on an instance level, it does not clarify contracts that are not behaviorally observable on an instance level (e.g., contracts that deal with privacy concerns, or with global properties of the predictions over a large sample set).

%move the old section to section_nine_archive.tex 
\section{Discussion} \label{sec:discussion}

The basis of our formalization of \HumanAI trust is the basic definition of interpersonal trust from sociology (\sect{sec:trust-basic-definition}). In this section, we first discuss additional aspects of interpersonal trust and human-machine trust (trust in automation), and their relevance to our definitions (\sect{sec:human_ai_definitions}). We then present two  directions for extension of our formalization inspired by other factors of interpersonal trust.

\subsection{Other Elements of Interpersonal Trust and Trust in Automation}
%\subsection{Elements of Trust in Social Science}

\noindentparagraph{\textbf{Trust vs Reliance.}} \label{subsec:reliance}

%A preliminary question we should ask is whether \HumanAI trust can even be considered trust from a perspective of sociology.
\citet{baier1986trust} introduces the term \textit{reliance} for trust in inanimate objects, such as trusting a stool to hold your weight. 
%\citet{baier1986trust} makes a distinction between trust and reliance, where reliance is considered to be a simpler variation of trust in inanimate objects, such as trusting a stool to hold your weight. %``I trust this stool to hold my weight.''
We may feel betrayed when our trust fails us, but this is not true for reliance because we did not attribute intent to the trustee \citep{tallant2020trust-commerce}. 
Despite the fact that AI models are tools, people anthropomorphize them and assign intent to them \citep{miller2017social, jacovi2020aligned-social}. This makes us believe that reliance is not suitable for AI models, and for this reason we aim to define \HumanAI \textit{trust} instead of reliance. This positions \HumanAI trust as a distinct sub-case of human-machine trust, or trust in automation, where otherwise the automation may not be attributed with intent (in which case, it is reliance). 
%As mentioned by \citet{miller2017social} and \citet{jacovi2020aligned-social}, humans do attribute human intent to AI models, so we elect to define human-AI trust as trust (rather than reliance), despite AI models being tools. However, more discussion in this area may be appropriate. 

\noindentparagraph{\textbf{Warranted and Justified Trust in Social Science.}} \label{subsubsec:warranted-trust-social}

Our definition of warranted \HumanAI trust (\sect{sec:human_ai_definitions}) %, as mentioned,
is trust that is caused by trustworthiness (to some contract). This definition is only applicable to \HumanAI trust, and is \textit{not} strictly relevant in sociology. 

Sociology elects to define trustworthiness and \textit{justified} trust by the effect, rather than the cause: \textbf{in interpersonal trust, the trustee is trustworthy, and the trust in them was justified, if the trust was not betrayed}. Two natural questions emerge: (1) why does sociology define interpersonal trust in this way? (2) and why do we not adopt this definition for \HumanAI trust?

The \textit{capability} of the trustee to maintain the trust, and their \textit{intent} to do so, are both prerequisites to interpersonal trust, \textit{but are not necessarily sufficient}. This is due to the complex nature of human interaction, with respect to elements of chance and outside factors (e.g., the difference between innocent mistake and negligence, `chains' of trust, and so on). This makes the assignment of blame difficult or ill-defined: e.g., if the trustee was fully capable of maintaining their trust, and intended to do so, but caught a common cold that ultimately prevented them from achieving their goal---it is difficult to define them as trustworthy or otherwise. 

\HumanAI trust does not share these limitations, as (1) the AI (by definition as an automation) does not possess real intent; and (2) the AI's capabilities are well-defined. As a result, we diverge from sociology definitions on trustworthiness and justified trust, and adopt a stricter causal definition of warranted trust.

\noindentparagraph{\textbf{Can Trust Become Warranted?}} In this work, we discuss warranted trust as a `state' that trust can become, on a binary. This is equivalent to the notion of trust becoming `calibrated' as discussed by \citet{lee2004trust-calibrated}. However, this is not realistic since trust can increase or decrease freely---on a sliding scale. But is the sliding scale, in itself, sufficient to describe trust?
% Naturally, this is not strictly true in reality, as trust can increase, decrease, or morph and change freely. Strong trust can very swiftly change to strong distrust, and vice-versa. 
More specifically, \citet{hoffman2017taxonomy-trust-automation} argues that trust cannot be positioned on a binary, or even a sliding scale, as trust is multi-dimensional. In his own words:
\begin{quote}
    % In domains such as information technology security and weather forecasting, workers can depend on dozens of software suites, each offering multiple functionalities, some suites running on the same computer, some running on different computers or servers, some using the same operating system, and some using different operating systems.
    In my own relation to my word processing software, I am positive that it will perform well in the crafting of simple documents, but I am simultaneously confident it will crash when the document gets long, or when it has multiple high-resolution images. And every time that there is a software upgrade, the trusting of many of the functions becomes tentative and skeptical. [Therefore,] trust is not a single state.
\end{quote}
We argue that \citeauthor{hoffman2017taxonomy-trust-automation}'s view is implicitly informed by a notion of contractual trust. In other words, \textbf{while general trust cannot be described by a single state (or sliding scale) of warranted or unwarranted, trust in a contract can,} since it is `calibrated' by the capability of the model to maintain the contract, where contracts are dimensions.

\subsection{Future Extensions of the Formalization}

\noindentparagraph{\textbf{Trust in the AI Developer.}}

% \todo{like ana said, this subsection can be presented as area for future work, rather than "what we're not doing". i wrote it like this for now...}
Literature in sociology often considers aspects of larger social constructs (beyond a single transaction of two people), specifically relationships and communities and the propagation of trust in them. A common theme in such models is the incentives of both parties (e.g., trust in family vs trust in a business partner). 

This notion is, of course, relevant to \HumanAI trust. For example, since there is a close relationship between workplaces with discriminatory practices and discriminatory AI tools \citep{west2019discriminating}, it is likely that those who are discriminated against have more incentive to trust AI tools produced by teams that represent them. By recognizing the incentives of the developer, the user may gain trust ``in the AI'', separately from the other causes described in this work.

AI as an automation does not embody intent. Formally, the intent of the AI developer manifests in the capability of the model to maintain specific contracts, rather than adopting some anthropomorphized notion of intent. Therefore, we make two claims on the nature of this trust.

\textbf{Trust in the AI model based on trust in the AI developer is an instance of \textit{interpersonal} trust by proxy, and not \HumanAI trust.} Therefore,  studies of the influence of the trustee's incentives on \HumanAI trust, should build on the existing research in sociology that investigates the influence of relationships and communities on interpersonal trust \cite{lewis1985trust}.
Consequently, \textbf{the question of whether this trust is \textit{warranted} or \textit{unwarranted} is ill-defined.} Our definition of warranted trust---as trust that is \textit{caused} by trustworthiness---is a definition that does not apply to interpersonal trust (as mentioned in \sect{subsubsec:warranted-trust-social}). 

To conclude, while trust in the AI developer \textit{could} possibly influence the user-AI interaction, and should therefore be studied and modeled, it is strictly not part of our model of \HumanAI trust. We therefore position this topic as future work towards a more nuanced model of trust in the interaction between people and AI.

\noindentparagraph{\textbf{Personal Attributes of the Trustor.}}

As mentioned, in this work we consider each interaction as a `clean slate' transaction of trust. Future work in this area may incorporate elements of the personal attributes of the trustor into the model, such as personality, socio-cultural background \cite{kopecka2020xai-cultural}, prior existence of trust or distrust, or the restoration of trust that has been betrayed.

% \subsection{Trust in Automation}

% \todo{i think we can delete this section, because we address trust in automation in various points over the work. but it's still worth putting a footnote somewhere just to acknowledge this area as a father or precursor to human-ai trust.}

% \noindentparagraph{Trust in automation.} The subject of trust in automation can be seen as a precursor to trust in artificial intelligence, and has been discussed at length before \citep{parasuraman1997automation-use-misuse,hoffman2017taxonomy-trust-automation}. While AI models have many peculiarities that are seldom discussed in ``trust in automation'' literature, in particular the differentiation from reliance (the attribution of intent to the model) and the lack of transparency (often black-box models), some progress in those areas is certainly relevant to us, which we will consider as well. %\ana{Is this specific to social sciences?} \todo{what do you mean?} \ana{This paragraph didn't perfectly fit the section that was titled `Trust in social sciences'.}

\section{Conclusion}

While Trust is a central component in the interaction between a user and the AI, current references to trust in the XAI literature are vague and under-specified. In this work, we provide several working definitions for trust-related terms, and answer the questions of what is necessary to allow trust to occur, and what is necessary for the goal of trust (anticipation of desired contracts) to be achieved. The goal of this formalization is to allow a disciplined method of developing trustworthy AI models that will incur trust in practice. We discuss intrinsic reasoning and extrinsic behavior as causes of warranted trust to be pursued in designing a trustworthy model. This is directly connected to XAI, which can provide the framework to verify whether a model is trustworthy or to create trustworthiness in the model. We further note that the question of trust in this context hinges on a notion of vulnerability, which is not satisfied by many evaluation methods currently used in XAI.

% \pagebreak

\noindentparagraph{\textbf{Takeaways.}} We collect the various conclusions in this work into guidelines that should inform the design of AI that is both trustworthy and trusted. 

% \pagebreak
% \newpage

\begin{enumerate}
    \item \textbf{The assessment of risk is necessary prior to the assessment of trust.} When deciding whether an AI requires trust, or when evaluating trust, verify the existence of vulnerability of the user in the actions of the AI, by: (1) verifying that the user considers some of the actions of the AI as unfavorable to them; and (2) verifying that the user considers both the favorable and unfavorable outcomes as realistic. 
    \item \textbf{AI developers should be explicit about the contracts that their models maintain.} AI developers should use the right affordances to make the relevant contracts explicit. This can help to avoid situations where a contract is implicitly assumed by a user, despite the developer not considering the contract to be upheld (unwarranted trust). 
    \item \textbf{Successful anticipation, while the goal of trust, is \textit{not} indicative of warranted trust.} The trust can be unwarranted if it is not sourced in trustworthiness, in which case, the anticipation may depend on a different variable that does not exist in other situations (such as the quality of the AI user interface). Therefore, though simulatability methods are useful and valuable as methods of assessing this property, it is dangerous to rely on them solely.
    % \item Static test-sets are a useful methodology to verify trustworthiness in specific contracts, or to accrue trust in those contracts. This is in contrast to the view that test-sets are compromised substitutes of deployment data.
    % \tm{The above seems to me to contradict what we say earlier in the paper? In fact, I would say that there is very little acceptance in the AI community that test sets are compromised. Accuracy on test sets is THE standard of most machine learning papers. The earlier part of the paper frames it as one method, but not sufficient on its own in many case. The above seems to argue the opposite??}
    \item \textbf{Trust is only ethically desirable if it is warranted.} Unwarranted trust is not guaranteed to accomplish its goal, since it is not sourced by trustworthiness. This can cause issues of abuse, disuse or misuse of the AI. While unwarranted trust may be in the interest of some parties, we assert that AI research should make efforts to both diagnose and avoid unwarranted trust by, among other things, identifying relevant contracts and assessing trustworthiness.
    \item \textbf{Distrust is not strictly undesirable if it is warranted.} Just as trustworthiness and warranted trust must both manifest for the AI to be useful in practice, so too must warranted distrust follow non-trustworthiness for contracts that are relevant to the application. Completely trustworthy AI, for all relevant contracts, may be prohibitively difficult to achieve---in which case warranted distrust is the mechanism that will allow the imperfect AI to be useful. 
    \item \textbf{Explanation seems to be uniquely positioned for \textsc{\small Human}-\textsc{\small AI} trust as a method for causing \textit{intrinsic} trust for general users.} Other causes of trust, such as empirical evaluation and authority, are extrinsic. This may help to explain the recent interest in XAI.
    \item \textbf{Methods of evaluating whether trust is warranted are underdeveloped, and require future work.} It is important to verify whether the trust of the user in an AI is warranted. However, there is currently little work on achieving this goal, and this questioned is positioned to be central for future research on \HumanAI trust.
    %\tm{Do you think there is something to add about XAI here? In particular, that perhpas the reasons there is such an  interest in explanation is because it is (uniquely?) positioned to be the only method that enabled intrinsic trust?} \todo{I think this could be nice, but didn't manage to think of any concrete wording. Maybe not necessary given the rest of our focus on XAI?}
\end{enumerate}

% \yg{what do we want / expect to happen due to this work, beyond people reading it? I think the last paragraph will be much stronger if it revolves around such expectations.} \todo{i added the paragraph below to try to answer this. i think it has some overlap with the takeaways afterwards, so i'm not sure it's necessary?}
% \yg{maybe move the "we hope" paragraph to after the takeaways?}
We hope that this work informs AI research in the following ways: (1) by encouraging more accurate discussion of trust in AI, through a more transparent definition of trust that includes the notions of contractual and warranted trust; (2) by encouraging claims on trust and on methods of causing or evaluating trust to be founded on, and distinguish between, the notions of intrinsic and extrinsic trust; (3) by requiring to explicitly recognize \emph{and verify} risk in user studies that make claims on trust.

\begin{acks}
We thank the anonymous reviewers for their helpful feedback. This project has received funding from the European Research Council (ERC) under the European Union's Horizon 2020 research and innovation programme, grant agreement No. 802774 (iEXTRACT) and from the Australian Research Council (ARC) Discovery Grant DP190103414: \emph{Explanation in Artificial Intelligence: A Human-Centred
Approach}. 
\end{acks}

% \begin{acks}
% \end{acks}

\bibliographystyle{ACM-Reference-Format}
\bibliography{sample-base}

% \appendix









\end{document}